\documentclass[sigconf]{acmart}
\usepackage{xspace}
\usepackage{colortbl}
\usepackage{xcolor}
\usepackage{balance}

\newenvironment{formal}
{\begin{quote}\em}
{\end{quote}}

\AtBeginDocument{%
  }

\setcopyright{acmlicensed}
\copyrightyear{2024}
\acmYear{2024}
\setcopyright{acmlicensed}\acmConference[CIKM '24]{Proceedings of the 33rd
ACM International Conference on Information and Knowledge
Management}{October 21--25, 2024}{Boise, ID, USA}
\acmBooktitle{Proceedings of the 33rd ACM International Conference on
Information and Knowledge Management (CIKM '24), October 21--25, 2024,
Boise, ID, USA}
\acmDOI{10.1145/3627673.3680020}
\acmISBN{979-8-4007-0436-9/24/10}

\newcommand{\sysname}{\texttt{LawLLM}\xspace}

\definecolor{customgreen}{HTML}{8ecfc9}
\definecolor{customyellow}{HTML}{ffbe7a}
\definecolor{customred}{HTML}{fa7f6f}
\definecolor{customblue}{HTML}{82b0d2}

\begin{document}

\title{LawLLM: Law Large Language Model for the US Legal System}

\author{Dong Shu}
\orcid{0009-0009-9785-3934}
\affiliation{%
  \institution{Northwestern University}
  \city{Evanston}
  \state{IL}
  \country{United States}}
\email{dongshu2024@u.northwestern.edu}

\author{Haoran Zhao}
\orcid{0000-0003-4563-9302}
\affiliation{%
  \institution{Northwestern University}
  \city{Evanston}
  \state{IL}
  \country{United States}
}
\email{haoranzhao2024@u.northwestern.edu}

\author{Xukun Liu}
\orcid{0000-0002-4608-9448}
\affiliation{%
  \institution{Northwestern University}
  \city{Evanston}
  \state{IL}
  \country{United States}
}
\email{xukunliu2025@u.northwestern.edu}

\author{David Demeter}
\orcid{0000-0001-7560-1132}
\affiliation{%
  \institution{Northwestern University}
  \city{Evanston}
  \state{IL}
  \country{United States}
}
\email{ddemeter@u.northwestern.edu}

\author{Mengnan Du}
\orcid{0000-0002-1614-6069}
\affiliation{%
  \institution{New Jersey Institute of Technology}
  \city{Newark}
  \state{NJ}
  \country{United States}
}
\email{mengnan.du@njit.edu}

\author{Yongfeng Zhang}
\orcid{0000-0003-2633-8555}
\affiliation{%
  \institution{Rutgers University}
  \city{New Brunswick}
  \state{NJ}
  \country{United States}
}
\email{yongfeng.zhang@rutgers.edu}

\renewcommand{\shortauthors}{Dong Shu et al.}

\begin{abstract}
In the rapidly evolving field of legal analytics, finding relevant cases and accurately predicting judicial outcomes are challenging because of the complexity of legal language, which often includes specialized terminology, complex syntax, and historical context. Moreover, the subtle distinctions between similar and precedent cases require a deep understanding of legal knowledge. Researchers often conflate these concepts, making it difficult to develop specialized techniques to effectively address these nuanced tasks. In this paper, we introduce the Law Large Language Model (\sysname), a multi-task model specifically designed for the US legal domain to address these challenges. \sysname excels at Similar Case Retrieval (SCR), Precedent Case Recommendation (PCR), and Legal Judgment Prediction (LJP). By clearly distinguishing between precedent and similar cases, we provide essential clarity, guiding future research in developing specialized strategies for these tasks. We propose customized data preprocessing techniques for each task that transform raw legal data into a trainable format. Furthermore, we also use techniques such as in-context learning (ICL) and advanced information retrieval methods in \sysname. The evaluation results demonstrate that \sysname consistently outperforms existing baselines in both zero-shot and few-shot scenarios, offering unparalleled multi-task capabilities and filling critical gaps in the legal domain. Code and data are available at \url{https://github.com/Tizzzzy/Law_LLM}.
\end{abstract}




\maketitle

\section{Introduction}
The development of Large Language Models (LLMs) has led to significant progress in computational linguistics, particularly impacting fields like legal analytics. Given the nature of legal language, which includes complex terminologies and context-specific logical frameworks, LLMs offer unprecedented capabilities in this domain \cite{naveed2023comprehensive}. The integration of LLMs into the legal field significantly boosts the efficiency of legal practitioners, such as lawyers and judges, by accurately interpreting their natural language input and generating most relevant responses. This reduces the need for extensive manual review of huge legal texts. Moreover, LLMs can provide lawyers with novel insights, revealing overlooked details and perspectives that can be critical in complex cases. Recent developments in legal domain have demonstrated the potential of LLMs in enhancing legal judgment predictions and handling various legal tasks effectively. For example, studies such as LM-CompEval-Legal \cite{shui2023comprehensive} have systematically evaluated the effectiveness of LLMs, other projects like PLJP \cite{wu2023precedent} and LoT \cite{jiang2023legal} have focused on integrating domain-specific models and advancing LLMs' understanding of legal reasoning.

Although these models have shown promise, there remain research challenges. First, these models generally address single-task challenges. In contrast, \sysname innovatively supports multiple legal tasks simultaneously, providing a more nuanced analysis of complex legal datasets and filling a critical void in the field. Second, another controversial area in the legal domain is the difference between precedent cases and similar cases \cite{qin2023incorporating}. Various models have been developed for precedent case recommendation, ranging from expert knowledge-based models to models based on natural language processing \cite{mentzingen2023automation, cao2024pilot, ma2023caseencoder, lewis2021precedent}. These approaches typically convert legal text into embeddings and calculate similarity at the embedding level, which aids in precedent selection. However, we believe that this approach is more on identifying similar cases with textual and contextual similarities, not precedent cases.

\begin{figure}[t]
  \centering
  \includegraphics[width=\linewidth]{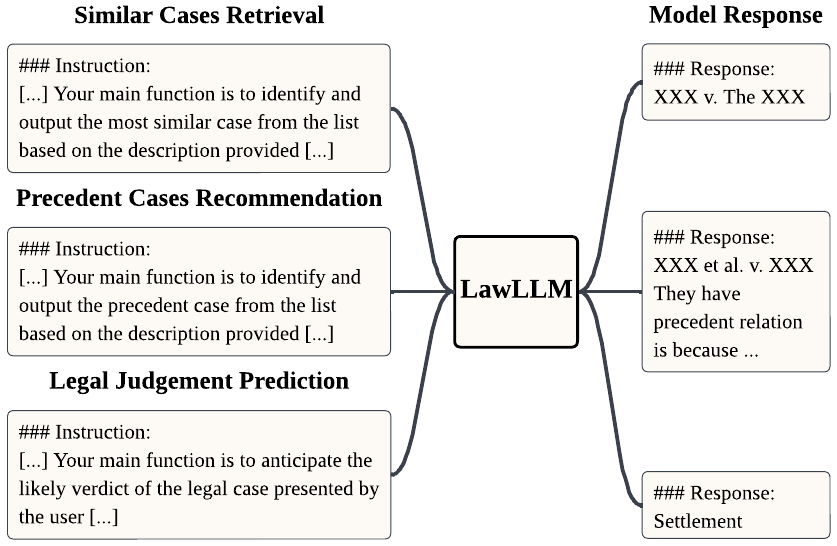}
  \caption{LawLLM supports three tasks: Similar Case Retrieval, Precedent Case Recommendation, and Legal Judgment Prediction.}
  \Description{overview of lawllm}
  \label{fig:intro}
\end{figure}

In our study, we emphasize the key differences between the two. Firstly, a precedent case must have been closed before the input legal case, ensuring its relevance and applicability to the current case under consideration. Secondly, precedent cases are those that were actually considered by judges in making their decisions, unlike similar cases which might not have been taken into account. Thirdly, similar cases share textual and thematic similarities in the case narrative or might fall into similar case categories, while precedent cases might seem unrelated at face value. It is also worth noting that while a legal case's precedent case can sometimes be the same as a similar case, this is not always the case. 

In this paper, we introduce the Law Large Language Model (\sysname), a multi-task LLM capable of Similar Case Retrieval (SCR), Precedent Case Recommendation (PCR) and Legal Judgment Prediction (LJP).
To build \sysname, we finetune Gemma-7B \cite{team2023gemini} using instruction tuning on United State real-life legal datasets and can perform LJP, PCR, and SCR tasks. The instructions for all three tasks fall under the classification category. By doing so, we can effectively minimize irrelevant and noisy options for the model, thereby improving its performance. We show the overall idea of \sysname in Figure \ref{fig:intro}. The development of \sysname also includes three innovative preprocessing approaches for transforming unstructured legal data into a trainable format. More specifically, for LJP, we summarize and extract verdicts from raw datasets and apply zero and few shot In-context learning (ICL) \cite{xie2021explanation, jin2024impact} technique to enhance the model performance. In PCR, \sysname connects ground truth legal cases' precedent relationships as a Knowledge Graph (KG), treating each case as a unique entity linked by precedential connections \cite{shu2024knowledge}. Additionally, the SCR task creates a legal case vector database and integrates advanced Information Retrieval (IR) techniques \cite{ethayarajh2019contextual, jin2024health}.

Our study presents \sysname as a pioneering model in the realm of legal LLMs. Our key contributions are given as follows:
\begin{itemize}
    \item We propose \sysname, which is adept at handling a range of legal tasks, including LJP, PCR, and SCR. This multi-task functionality is important in addressing the diverse requirements of the legal domain.
        
    
    \item \sysname distinguishes between precedent cases and similar cases, providing clarity on the objectives of each task. This clarification enables the future research to develop tailored strategies for those tasks.

    \item Experimental results indicate that \sysname outperformed all baseline models, including the GPT-4 model, across all three tasks. These results highlight \sysname's robust capabilities in the legal domain. 

\end{itemize}


\begin{figure*}
  \includegraphics[width=\textwidth]{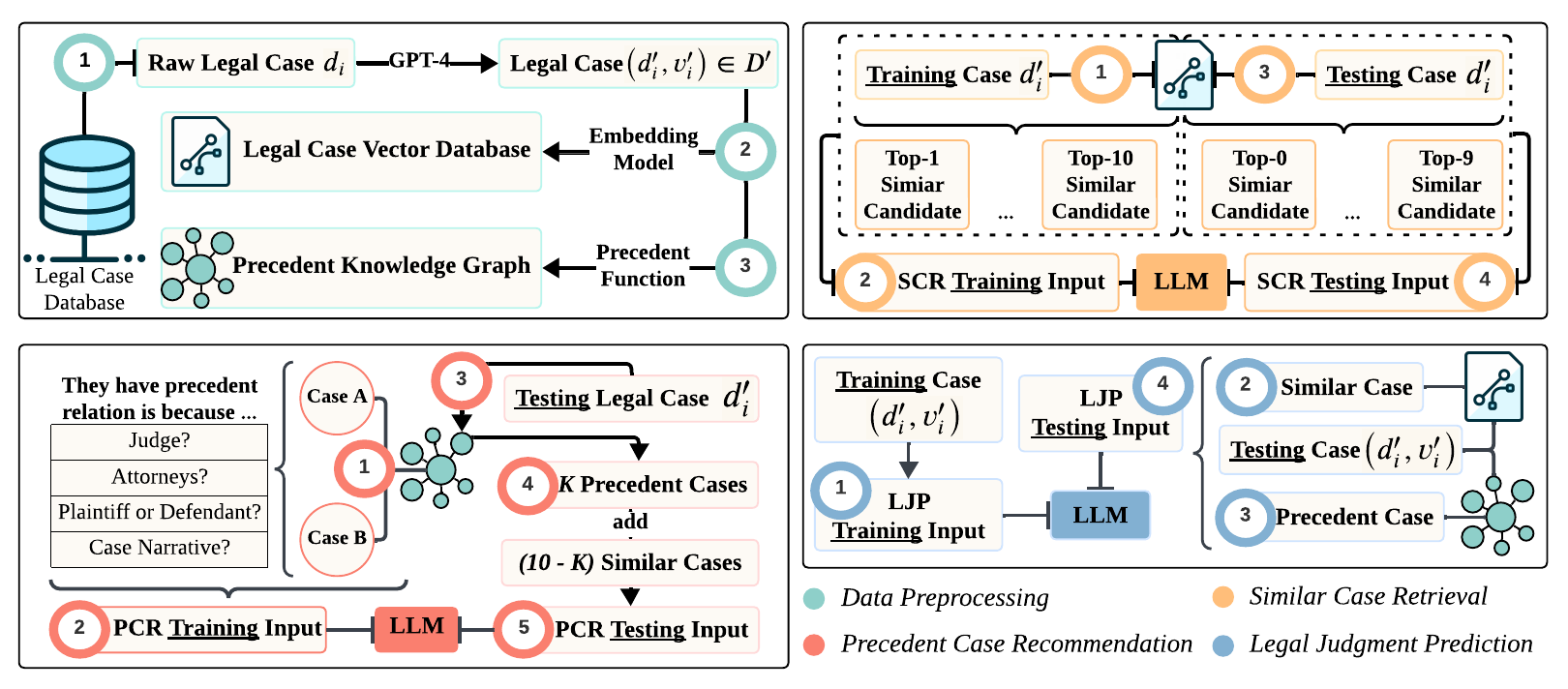}
  \caption{An overview of our LawLLM: Data Preprocessing is in the upper left in \textcolor{customgreen}{green}, Similar Case Retrieval Processing is in the upper right in \textcolor{customyellow}{yellow}, Precedent Case Recommendation is in the lower left in \textcolor{customred}{red}, and Legal Judgment Prediction is in the lower right in \textcolor{customblue}{blue}.}
  \Description{work overflow of lawllm}
  \label{fig:overflow}
\end{figure*}

\section{Related Work}

Legal AI is significantly increasing the efficiency and effectiveness of the legal community. AI technologies, specifically Large Language Models (LLMs), are leading the way in automating complex tasks like document analysis, case prediction, and legal research \cite{zhong2020does, wu-etal-2023-precedent}. LLMs utilize advanced algorithms and data analytics to process and generate legal texts, which leads to significant improvements in speed and accuracy \cite{zhou2023boosting}. In this section, we introduce the various applications of Legal AI and LLMs in legal practices.

\subsection{Precedent Case Recommendation}
The recommendation of precedent cases is a fundamental aspect of legal practice, as previous verdicts significantly affect current legal decisions. The field has evolved from early keyword-based searches and manual annotations to more complicated AI-driven models that improve retrieval efficiency and contextuality. \citet{wu-etal-2023-precedent} proposed the Precedent-Enhanced Legal Judgment Prediction framework, which combines LLMs with domain-specific expertise to improve legal prediction accuracy significantly. \citet{ma2023incorporating} developed the Structured Legal Case Retrieval system, which uses structural information from legal documents to improve case search precision and contextual relevance. Moreover, \citet{su2023caseformer} proposed the Caseformer. This innovative pre-training framework learns from a vast corpus of legal texts to refine case retrieval and contextualization across multiple languages.

\subsection{Similar Case Retrieval}
Besides precedent recommendation, retrieving similar cases, those sharing analogous facts or legal issues, is crucial for comprehensive legal analysis and strategy formulation. Traditionally, this process required extensive manual labor, with professionals needing to dig through large case databases \cite{mandal2021unsupervised, ma2023incorporating}. Today advances in NLP and machine learning have transformed this task, allowing semantic content extraction and comparison across documents. \citet{kang2013retrieval} enhanced similarity-based retrieval by incorporating associative knowledge. This approach refines retrieval outcomes by leveraging similarity and associative analyses, a technique also proven effective in other fields such as medical diagnosis and IT service management. \citet{mandal2021unsupervised} analyzed textual similarity techniques on an Indian Supreme Court dataset and discovered that traditional methods like TF-IDF outperform modern context-aware models like BERT. \citet{wu2021joint} studied semantic retrieval in the Chinese judicial system and developed a model that generates knowledge graphs for cases to improve trial accuracy and fairness. These technological advances have greatly simplified legal research, making it more effective and comprehensive.

\subsection{Legal Judgment Prediction}
Predicting legal judgments involves estimating potential verdicts based on a deep analysis of historical data and established legal standards. Initial models in this field were relatively simple, mainly depending on linear algorithms incapable of capturing the various aspects of legal reasoning. \citet{wang2020study} CNN-BiGRU multi-task learning model improves prediction accuracy through the utilization of shared information from related legal subtasks. \citet{chalkidis2019neural} used European Court of Human Rights data to establish robust performance benchmarks for long legal texts using hierarchical BERT. \citet{rusnachenko2023nclu_team} showed attention-based methods could improve system performance by optimizing document preprocessing and attention mechanisms in competition contexts. These models predict outcomes and are constantly learning from new cases to improve their accuracy, demonstrating the adaptability of LLMs in legal judgment prediction.

\subsection{LLMs in the Legal Domain}
Prior to the development of large language models (LLMs), pre-trained language models (PLMs) for specific domains were explored, such as Lawformer, which is to process lengthy Chinese legal documents using a Longformer-based architecture \cite{xiao2021lawformer}. Researchers discovered that models like GPT-4 could successfully pass bar exams as LLMs gained attention, demonstrating profound abilities in legal reasoning and text generation \cite{katz2024gpt}. This success resulted in the growth of legal domain-specific LLMs, such as Chatlaw, which utilizes conversational AI to improve user interactions with legal systems \cite{cui2023chatlaw}. In this vein, SaulLM-7B was introduced as the first LLM explicitly designed for comprehending and generating legal texts, leveraging a substantial legal corpus to achieve state-of-the-art performance \cite{colombo2024saullm}. LLMs' influence extends beyond specific tasks to broader legal operations. These applications range from document automation, where LLMs assist in drafting and reviewing legal documents, to compliance monitoring, which ensures adherence to regulatory standards \cite{Sun2023A}. LLMs simplify complex legal processes for non-specialists and lower barriers to legal advice \cite{Goodson2023Intention}. This broad application of LLMs demonstrates their broad application and the potential for continued innovation in the legal sector.

Despite the success of those contemporary works, these models primarily focus on utilizing LLMs' understanding and capabilities to perform general legal question answering. However, \sysname is designed to leverage the LLMs' comprehension and learned abilities to predict and perform specific tasks within the legal domain.

\section{Methodology}

In this study, we propose the Law Large Language Model (\sysname) to address three critical tasks within the legal domain: Similar Case Retrieval (SCR), Precedent Case Recommendation (PCR), and Legal Judgment Prediction (LJP). Our methodological framework, illustrated in Figure \ref{fig:overflow}, is divided into four distinct parts: Data Preprocessing, SCR Processing, PCR Processing, and LJP Processing.

\subsection{Data Preprocessing}

Our approach begins with the systematic collection of case data from legal databases, denoted as \( \mathcal{D} \). We make sure all collected raw case data, \( d_i \in \mathcal{D} \), encompasses a variety of information as below:

\begin{equation}
\begin{aligned}
d_i = \{& \text{Title, } \text{Date, } \text{Judge, } \text{Plaintiff(s), } \text{Plaintiff's Attorney(s), }\\ 
& \text{Defendant(s), }\text{Defendant's Attorney(s), } \text{Case Detail, } \\
& \text{Precedent Relationship}\}.
\end{aligned}
\end{equation}

As depicted in the upper left of Figure \ref{fig:overflow}, data preprocessing consists of three primary steps:

\vspace{3pt}\noindent\textbf{Step 1.}\, Given the voluminous nature of the textual content within case detail and their often implicit verdicts, we utilize a GPT-4 \cite{achiam2023gpt} model to extract core information and summarize each case. This step reduces information overload and ensures the adaptability of our dataset to the constraints of Gemma, particularly with token size limitations. The GPT-4 preprocess instruction is shown here: 

\begin{formal}
I have a legal case description and require two distinct pieces of information:
\newline
1. Summary: Please provide a detailed summary of the case, focusing on
the facts and events. Exclude any information about the verdict.
\newline
2. Verdict: State the verdict of the case, consider the following categories:
\newline
- Plaintiff win\newline
- Defendant win\newline
- Settlement\newline
- Case dismissal\newline
- Unsure\newline
If the verdict is mentioned, respond exclusively with the chosen
categories ONLY. If the outcome is not explicitly mentioned or cannot
be inferred from the information given, please respond with `unsure'
only.
\newline
Format your responses as follows:\newline
\# - For the summary, begin with `Answer 1:'\newline
\# - For the verdict, start with `Answer 2:'\newline
Here is the description of the case:\newline
[Case Description...]
\end{formal}

The output of this step includes a summarized case and a labeled verdict, formatted as follows:

\begin{equation}
\begin{aligned}
\text{Case Summary, Verdict} = \text{LLM}(&\text{Case Detail}, \\
& \text{Maximum Token} \mid d_i).
\end{aligned}
\end{equation}

For each legal case \(d_i\), we reorganize the data into a new format \(d_i'\), defined as:

\begin{equation}
\begin{aligned}
d_i' = \{& \text{Title, } \text{Date, } \text{Judge, } \text{Plaintiff(s), } \text{Plaintiff's Attorney(s), }\\ 
& \text{Defendant(s), }\text{Defendant's Attorney(s), } \\
& \text{Case Summary}\}.
\end{aligned}
\end{equation}

\begin{equation}
\begin{aligned}
D' = \left\{ \left( d_1', v_1' \right), \left( d_2', v_2' \right), \ldots, \left( d_n', v_n' \right) \right\}.
\end{aligned}
\end{equation}

There are some constraints when we separate the \( D' \) into training and testing data. We make sure that all legal cases have at least five precedent relationships. To ensure a balance training, the training dataset has 25\% from each of the following categories: plaintiff wins, defendant wins, settlements, and case dismissals. We also make sure that all testing legal cases have at least five precedent relationships connect to the training dataset, further explanation is given in Section \ref{data_splits} Data Splits.

\vspace{3pt}\noindent\textbf{Step 2.}\, After Step 1, all training legal cases \( d_i' \) are transformed into high-dimensional vectors using the OpenAI Embedding model. This vector database is later used to retrieve the top-$k$ similar cases based on semantic and contextual similarities.

\vspace{3pt}\noindent\textbf{Step 3.}\, This step involves converting the precedent case relationships from our training dataset into a knowledge graph (KG). Defined as $KG = (E, R, L)$, where $E$ represents entities, $R$ represents binary relationships (indicative of precedent relations), and $L \subseteq E \times R \times E$ represents the set of triples forming the graph's edges. Each triple $(e_s, r, e_t) \in L$ indicates a directed edge from source entity $e_s$ to target entity $e_t$ via relationship $r$. The KG data structure simplifies the complex task of identifying relevant precedent cases, turning it into a entity prediction problem, i.e., given a query of $(e_s, r, ?)$, the model will predict the missing entity.

We further customize data processing for SCR, PCR, and LJP tasks, ensuring a robust and effective implementation of \sysname.

\subsection{Similar Case Retrieval}

As depicted in the upper right of Figure \ref{fig:overflow}, the SCR process is divided into two phases: training (Steps 1-2) and testing (Steps 3-4).

\vspace{3pt}\noindent\textbf{Training Phase.}\, During training, each training case \(d_i' \) is inputted into the vector database, which generates the top 10 candidate cases. These cases are then randomized in order and formulated into the SCR training instruction. Here is an example SCR model input:

\begin{formal}
\#\#\# Instruction:\newline
You are a legal expert who specializes in comparing user-supplied legal cases to a list of candidate legal cases, which includes titles and content. Your main function is to identify and output the title of the most similar case from the list based on the description provided. \newline
You should only output the case title and not any other information. \newline
Consider the following choices:\newline Choice 1: \newline [Case 1...] \newline Choice 2: \newline ... \newline  Choice 10: \newline [Case 10...]\newline
\#\#\# Input:\newline
[Input Case...]
\end{formal}

In this scenario, the SCR task instruction will fall into the classification category, which provides the model with 10 cases to choose the most similar one. It is important to note that the top-0 similar case is the case $d_i'$ itself, so in practice, we retrieve the top-1 to top-10 similar cases from the vector database, and the top-1 case from this selection serves as the ground truth for this training task.

\vspace{3pt}\noindent\textbf{Testing Phase.}\, 
The testing phase mirrors the training process as we initially retrieve the top 10 similar cases from the vector database. However, during testing, we retrieve cases ranked from top-0 to top-9, as the test case itself is not included in the vector database. The model's expected response depends on the evaluation metrics we use: top-1, top-3, and top-5. For the top-1 metric, we expect \sysname to identify the most similar case as the top result. The top-3 metric evaluates whether the model's answer falls within the top three retrieved candidates, while the top-5 metric extends this evaluation to include the top five candidates. 

\subsection{Precedent Case Recommendation}

The Precedent Case Recommendation (PCR) within \sysname utilizes a unique approach by employing a precedent case knowledge graph (KG), which differentiates itself from conventional PCR methods that often speculate on potential precedent relationships. Our system instead relies on confirmed precedent pairs, as illustrated in the lower left of Figure \ref{fig:overflow}, where Steps 1 and 2 constitute the training phase and Steps 3-5 are the testing phase.

\vspace{3pt}\noindent\textbf{Training Phase.}\, From the previously established KG, for each confirmed triple $(e_s, r, e_t)$, we utilize BERT embeddings \cite{devlin2018bert} to evaluate the similarity between various case features (e.g., Judge, Case Detail, Plaintiff, or Defendant), denoted as \{\( F_1, F_2, ... F_j\)\}. We calculate the similarity score \( S_i \) for each feature pair \( F1_i \) and \( F2_i \), as follows:

\begin{equation}
\label{eq:1}
\begin{aligned}
S_{i} = \text{sim}(BERT(F1_{i}), BERT(F2_{i})), &\\ \quad i \in \{1-j\}
\end{aligned}
\end{equation}

The highest similarity score across all features determines the primary factor underlying their precedent relationship:

\begin{equation}
\label{eq:2}
\begin{aligned}
\text{Primary Factor} = \max(S_{1}, S_{2}, ..., S_{j}).
\end{aligned}
\end{equation}

During the training input creation, we present a total of 10 choices for the model. The ground truth precedent case $e_t$ is randomly placed among these choices, with the other 9 selections filled with similar, yet non-precedent, cases from the vector database. This setup aims to teach the model that textual similarity does not necessarily imply a precedent relationship. The model's expected output includes the correct precedent case $e_t$ and the reasoning for its selection (i.e, which primary factor caused this precedent relationship). An example of the model input is:

\begin{formal}
\#\#\# Instruction:\newline
You are a legal expert who specializes in comparing user-supplied legal cases to a list of candidate legal cases, which includes titles and content. Your main function is to identify and output the precedent case from the list based on the description provided. \newline
You should only output the reasoning process and case title. \newline
Consider the following choices:\newline Choice 1: \newline [Case 1...] \newline Choice 2: \newline ... \newline  Choice 10: \newline [Case 10...]\newline
\#\#\# Input:\newline
[Input Case...]
\end{formal}

\vspace{3pt}\noindent\textbf{Testing Phase.}\, For each test case, since we made sure there are at least five precedent cases in the training dataset, we can identify $k$ precedent cases from the KG (which structured by the training dataset) as ground truths, where $k$ aligns with the top-$k$ evaluation metrics. For the top-1 metric, a single ground truth precedent case is selected, while for top-3 and top-5 metrics, 3 and 5 ground truths are selected, respectively. The remaining slots of $10 - k$ are filled with similar cases. The model is then tasked with selecting one of $k$ precedent cases and explaining the reasoning behind its choice.

\subsection{Legal Judgment Prediction}

The Legal Judgment Prediction (LJP) processing utilizes the dataset \(D'\) constructed during the data preprocessing stage. This dataset pairs each processed legal case \(d_i'\) with its corresponding verdict \(v_i'\). As illustrated in the lower right of Figure \ref{fig:overflow}, the training phase involves step 1 and the testing phase involves rest of steps.

\vspace{3pt}\noindent\textbf{Training Phase.}\, We use $\left( d_i', v_i' \right)$ to establish a four-category classification training input, Plaintiff wins, Defendant wins, Settlement, or Case Dismissal. Each case's corresponding verdict \(v_i'\) serves as the label for training. Here is an example of the model input:

\begin{formal}
\#\#\# Instruction:\newline
You are a legal expert who specializes in predicting outcomes for legal cases. Utilize your internal knowledge base to predict verdict. Your main function is to anticipate the likely verdict of the legal case presented by the user.
\newline
You should only output the verdict and not any other information. \newline
Consider the following choices:\newline 1. Defendant Wins \newline 2. Plaintiff Wins \newline 3. Settlement \newline 4. Case Dismissal\newline
\#\#\# Input:\newline
[Input Case...]
\end{formal}

\vspace{3pt}\noindent\textbf{Testing Phase.}\, During the testing phase, we evaluate \sysname with both zero-shot and few-shot in-context learning (ICL) scenario. In few-shot ICL, we enhance each test case \(d_i'\) with additional contextual information, one similar case and one precedent case. Its precedent cases is sourced from our KG, and one is randomly selected to be included in the test input. Simultaneously, a most similar case is retrieved from the vector database. This approach ensures that the model's predictions are influenced by relevant legal precedents and similar case facts, thereby improving the accuracy and reliability of the judgment predictions.

\subsection{Unified Model Fine-Tuning}

Our methodology involves a unified fine-tuning strategy for the \sysname, leveraging a combined dataset with three tasks. This dataset, denoted as \(\text{Dataset}_{\text{combined}} = \text{LJP} \oplus \text{PCR} \oplus \text{SCR} \). We employ a cutting-edge 4-bit quantized Low-Rank Adaptation (LoRA) technique to instruction fine-tune the Gemma model. We use the cross-entropy loss function $L$ during the LoRA. It calculates the difference between the model's predicted token probabilities and the actual token probabilities in the expected output sequence. In the following equation, $n$ represents the length of the expected output sequence, $x$ represents the input instruction, and $y_i$ denotes the i-th token in the expected output sequence.

\begin{equation}
\label{eq:1}
L = -\sum_{i=1}^{n} \log P(y_i | x, y_1, y_2, ..., y_{i-1}).
\end{equation}

\section{Experiments}

In this section, we conduct experiments to evaluate the performance of \sysname on three tasks: Similar Case Retrieval (SCR), Precedent Case Recommendation (PCR), and Legal Judgment Prediction (LJP).

\subsection{Experimental Setup}

\vspace{3pt}\noindent\textbf{Datasets.}\,
We conduct our experiments on the CaseLaw dataset, initiated by Harvard Law School's Library Innovation Lab as part of the CaseLaw Project \cite{caselawproject}. This database encompasses a wide range of court cases from both state and federal in the United States. The project primarily focuses on democratizing access to American legal information, particularly through its Case Access Project (CAP), which is aimed at providing free and public access. The statistics of the CaseLaw dataset used in our experiments are shown in Table \ref{tab:dataset}.

\begin{table}[h]
\centering
\caption{Datasets Statistics}
\label{tab:dataset}
\begin{tabular}{ll}
\hline
DATASETS                         & CaseLaw \\ \hline
Language                         & English \\
\# State and Federal Totals      & 6,930,777 \\
\# Train case                    & 1,00,000  \\
\# Test case                     & 200,000   \\
Avg. length per case (words)     & 2695.38 \\
\hline
\end{tabular}
\end{table}

\vspace{3pt}\noindent\textbf{Evaluation Metrics.}\,
As previously mentioned, we employ top-$k$ metrics to evaluate the performance of the SCR and PCR tasks. Specifically, we use top-1, top-3, and top-5 metrics. These metrics measure the model's precision in identifying the correct response from a pool of 10 choices. For example, the top-1 metric requires the model to return the top choice as the answer. The top-3 and top-5 metrics provide more flexibility, allowing the correct answer to be anywhere within the top three or top five choices, respectively.

In addition to top-$k$ metrics, we evaluate the hallucination rate of models using a `not-found' metric. This metric tracks the proportion of responses that are entirely fabricated and do not match any of the 10 given choices. By measuring the `not-found' rate, we aim to understand how often models produce answers unrelated to the provided options, offering insight into their reliability.

For the LJP task, we employ accuracy and F1-score \cite{sai2022survey} metrics to gauge the model's performance. Accuracy calculates the proportion of correctly predicted verdicts across all cases, providing a direct measure of overall prediction performance. The F1 score ranging from 0 to 1, combines precision and recall into a single harmonic mean, offering a balanced evaluation of the model's effectiveness.


\vspace{3pt}\noindent\textbf{Data Splits.}\,
\label{data_splits}
As previously noted, our data are split according to three constraints. 
\begin{itemize}
    \item \textbf{Constraint 1:} For PCR, we employ top-$k$ evaluation metrics, which means each case has to have a minimum of five precedent cases, allowing us to identify $k$ ground truths.

    \item \textbf{Constraint 2:} We must ensure that when a test case is evaluated, its ground truth precedent case can be located within the knowledge graph formed by the training cases. Therefore, each test case must have at least five precedent cases present in the training data. 
    \item \textbf{Constraint 3:} To ensure balanced model training for Legal Judgment Prediction (LJP), the training data's verdict distribution should consist of 25\% for each possible outcome: plaintiff wins, defendant wins, settlements, and dismissals.

\end{itemize}

These approaches result in a total of 1,000,000 cases for training and 200,000 cases for testing.

\vspace{3pt}\noindent\textbf{Comparing Baselines.}\,
Our model is evaluated against advanced baselines including LLaMa2-7b \cite{touvron2023llama}, Gemma-7b \cite{team2023gemini}, Vicuna-13b \cite{zheng2024judging} and Guanaco-13b \cite{dettmers2024qlora}, alongside the larger and more advanced GPT-3.5 and GPT-4 models \cite{achiam2023gpt}. Each model undergoes the same testing phase to ensure a consistent and fair comparison of their multi-task capabilities within the legal domain.

\vspace{3pt}\noindent\textbf{Implementation Details.}\,
We conducted the training of our model over 10 epochs using an A40 GPU. To ensure compatibility, we monitored the input token size, capping it at 4096 tokens to align with Gemma's maximum token capacity. Additionally, we configured the model's dropout rate at 0.1 and set the learning rate to $2e^{-4}$.

\subsection{Similar Case Retrieval Results}

According to Table \ref{table:scr}, \sysname outperformed the baseline models in all categories. Specifically, it achieved the highest accuracy in top-1, top-3, and top-5 retrieval rates, with scores of 29.8\%, 63.2\%, and 81.6\% respectively. Remarkably, it also demonstrated minimal hallucination, as indicated by the not-found rate of 0.1\%.

\begin{table}[h]
\centering
\caption{SCR Test Results}
\begin{tabular}{lcccc}
\toprule
\midrule
Method          & top-1 $\uparrow$ & top-3 $\uparrow$ & top-5 $\uparrow$ & Not Found $\downarrow$ \\
\midrule
llama2-7b       & 0.083   & 0.197   & 0.309   & 0.406 \\
gemma-7b        & 0.181   & 0.428   & 0.536   & 0.121 \\
vicuna-13b      & 0.185   & 0.372   & 0.564   & 0.187 \\
guanaco-13b     & 0.077   & 0.214   & 0.375   & 0.372 \\
gpt3.5          & 0.219   & 0.579   & 0.691   & 0.148 \\
gpt4            & 0.274   & 0.526   & 0.708   & 0.005 \\
\rowcolor{gray!40} \sysname & \textbf{0.298}   & \textbf{0.632}   & \textbf{0.816}   & \textbf{0.001} \\
\midrule
\bottomrule
\label{table:scr}
\end{tabular}
\end{table}

Comparatively, GPT-4 showed strong performance with top-1, top-3, and top-5 accuracies of 27.5\%, 52.5\%, and 70.5\%, and a low not-found rate of 0.5\%. GPT-3.5 also performed well, especially in the top-3 and top-5 metrics. On the other hand, models like LLaMa2-7b and Guanaco-13b displayed higher not-found rates, indicating a tendency towards hallucination.

The results underscore the effectiveness of our \sysname model in accurately retrieving similar cases while minimizing the risk of generating irrelevant or nonexistent cases.

\subsection{Precedent Case Recommendation Results}

According to Table \ref{table:pcr}, the \sysname model again outperformed other baseline methods. It achieved the best results with a top-1 rate of 31.8\%, top-3 rate of 59.7\%, and top-5 rate of 83.2\%. Additionally, the \sysname model exhibited an low not-found rate of 0.1\%.

\begin{table}[h]
\caption{PCR Test Results}
\begin{tabular}{lcccc}
\toprule
\midrule
Method          & top-1 $\uparrow$ & top-3 $\uparrow$ & top-5 $\uparrow$ & Not Found $\downarrow$ \\
\midrule
llama2-7b        & 0.069   & 0.148   & 0.343   & 0.479 \\
gemma-7b         & 0.187   & 0.386   & 0.519   & 0.124 \\
vicuna-13b       & 0.175   & 0.352   & 0.506   & 0.203    \\
guanaco-13b      & 0.073   & 0.198   & 0.357   & 0.383    \\
gpt3.5           & 0.154   & 0.325   & 0.504   & 0.165 \\
gpt4             & 0.262   & 0.514   & 0.697   & 0.007 \\
\rowcolor{gray!40} \sysname & \textbf{0.318}   & \textbf{0.597}   & \textbf{0.832}   & \textbf{0.001} \\
\midrule
\bottomrule
\label{table:pcr}
\end{tabular}
\end{table}

Among the baseline models, GPT-4 was a strong performer, with high accuracy in top-1, top-3, and top-5 metrics, alongside a very low not-found rate, suggesting reliable and accurate recommendations. In contrast, models like LLaMa2-7b and Guanaco-13b showed higher not-found rates, highlighting challenges in providing relevant case recommendations. The overall results demonstrate the effectiveness of the \sysname model in PCR task, outstripping baseline models in both accuracy and reliability.

One notable insight from comparing SCR and PCR results is that most baseline models exhibited a performance drop in the PCR task compared to SCR. For instance, the GPT-4 model achieved scores of 27.4\%, 52.6\%, 70.8\%, 0.5\% in SCR top-$k$ and ``Not Found'' metrics, while in the PCR task, its scores dropped to 26.2\%, 51.4\%, 69.7\% and 0.7\%. This decline underscores the greater difficulty of identifying precedent cases compared to similar cases, as models cannot rely solely on textual similarity when determining precedent relationships. Instead, they must consider nuanced factors such as legal relevance. This performance difference reinforces the our previous assertion that precedent cases are distinct from similar cases, emphasizing the importance of distinguishing between the two concepts in the legal domain.

We conducted an analysis to identify the factors that are predominantly considered by \sysname when determining a precedent relationship under the top-1, top-3, and top-5 settings. This analysis involves comparing the frequency with which each factor is chosen as the primary determinant in our model against the ground truth (GT) distribution. As shown in Table \ref{table:factor}, the GT distribution is heavily weighted towards the `Case Detail' factor, with some toward other factors. In the top-1 scenario, where there is only one correct precedent case among nine similar cases, our model strongly focuses on the `Case Detail' factor. This bias likely stems from the GT distribution's heavy emphasis on `Case Detail,' leading our model to prioritize this factor, especially when faced with numerous similar cases that serve as potential distractions. However, as the pool of correct answers expands to three and five in the top-3 and top-5 scenarios respectively, \sysname begins to diversify its focus slightly to include other factors, although `Case Detail' continues to dominate. This trend indicates a move towards a more balanced approach in factor consideration as the number of correct choices increases, suggesting that \sysname adjusts its focus based on the availability of correct answers, while still reflecting the main emphasis observed in the ground truth data.

\begin{table}[h]
\caption{Primary Factor Percentage Comparison}
\begin{tabular}{lcccc}
\toprule
\midrule
Factor          & LawLLM    & LawLLM    & LawLLM    & GT   \\
                & top-1     & top-3     & top-5     &       \\
\midrule
Title           & 0.000   & 0.000   & 0.000   & 0.000    \\
Date            & 0.000   & 0.000   & 0.000   & 0.000     \\
Judge           & 0.027   & 0.054   & 0.116   & 0.149     \\
Plantiff(s)     & 0.002   & 0.009   & 0.013   & 0.027     \\
Defendent(s)    & 0.004   & 0.012   & 0.025   & 0.041     \\
Case Detail     & 0.967   & 0.925   & 0.846   & 0.783     \\
\midrule
\bottomrule
\label{table:factor}
\end{tabular}
\end{table}

\subsection{Legal Judgment Prediction Results}

As shown in Table \ref{table:ljp}, the \sysname surpasses all baseline methods in both zero-shot and few-shot scenarios for the LJP task. In the zero-shot scenario, LawLLM achieves an accuracy of 0.636 and an F1 score of 0.591, significantly outperforming the second best model, GPT-4, which scores 0.573 and 0.563 in accuracy and F1, respectively. In the few-shot scenario, LawLLM maintains its superior performance, reaching an accuracy of 0.794 and an F1 score of 0.758. These results show a considerable improvement over GPT-4, the closest competitor, which scores 0.732 in accuracy and 0.712 in F1.

Additionally, all models demonstrate higher performance in the few-shot in-context learning (ICL) scenario compared to the zero-shot setting. For instance, LLaMA2-7b shows an increase from 0.235 to 0.473 in accuracy, and from 0.239 to 0.455 in F1 score. This pattern indicates that all models benefit from incorporating a few ICL examples, which helps them better understand the task.



\begin{table}[h]
\centering
\caption{LJP Test Results}
\begin{tabular}{lcccc}
\toprule
\midrule
Method          & Accuracy $\uparrow$ & F1 $\uparrow$ & Accuracy $\uparrow$ & F1 $\uparrow$ \\
                & (Zero-shot) & (Zero-shot)  & (Few-shot) & (Few-shot) \\
\midrule
llama2-7b        & 0.235   & 0.239   & 0.473   & 0.455 \\
gemma-7b         & 0.317   & 0.287   & 0.568   & 0.527 \\
vicuna-13b       & 0.503   & 0.432   & 0.645   & 0.594    \\
guanaco-13b      & 0.281   & 0.247   & 0.491   & 0.463    \\
gpt3.5           & 0.558   & 0.546   & 0.679   & 0.647 \\
gpt4             & 0.573   & 0.563   & 0.732   & 0.712 \\
\rowcolor{gray!40} \sysname & \textbf{0.636}   & \textbf{0.591}   & \textbf{0.794}   & \textbf{0.758} \\
\midrule
\bottomrule
\label{table:ljp}
\end{tabular}
\end{table}

\section{Conclusions and Future Work}

In this study, we introduced the Law Large Language Model (\sysname), a multi-task LLM specifically designed for the US legal domain. By leveraging unique data processing techniques tailored for each task, \sysname effectively handles Similar Case Retrieval (SCR), Precedent Case Recommendation (PCR), and Legal Judgment Prediction (LJP). Furthermore, we emphasized the crucial distinctions between precedent relationships and textual similarity, providing insights that can inform future research in developing task-specific models. Our results consistently demonstrated that \sysname outperforms existing baseline models, showcasing its superior multi-task capabilities.

In the future, we aim to expand the scope of \sysname by incorporating additional legal tasks to further enhance its versatility and practical applicability. This will involve exploring emerging challenges in legal analytics and integrating new datasets that reflect diverse legal contexts. Moreover, we plan to refine our data processing techniques and in-context learning methodologies to improve the model's understanding of legal nuances and precedents. 

\bibliographystyle{ACM-Reference-Format}
\balance
\bibliography{sample-base}

\newpage

\appendix
\section{Appendix}

\subsection{The Choice of 10}
\label{appendix:10}

According to Table \ref{table:10choice}, we have several reasons for giving 10 choices to the model in the Similar Case Retrieval (SCR) and Precedent Case Recommendation (PCR) tasks:

\begin{itemize}
    \item Given that we utilize top-$k$ evaluation metrics where $k$ is 1, 3, and 5, the number of choices must be greater than 5.
    \item Our exploration revealed that when the number of choices exceeds 11, some inputs surpass the maximum token length of the Gemma model. Thus, the appropriate range for the number of choices lies between 6 and 11.
    \item We randomly selected 1,000 test cases to evaluate the performance of \sysname. From Table \ref{table:10choice}, we observe that with 6 or 7 choices, the model’s performance in top-5 metric approaches 100\% due to the limited challenge of smaller sets. Therefore, 6 or 7 choices are not the optimal option. Also, ``Not Found'' cases only emerge when the choice size reaches 9 in PCR tasks and 10 in SCR tasks. Ultimately, we chose 10 as the optimal size, as the model perform similarly, and it provides more challenge to the model.

\end{itemize}
    
\begin{table}[h]
\centering
\caption{Choice Size Results}
\begin{tabular}{lcccc}
\toprule
\midrule
Method          & top-1    & top-3    & top-5    & Not Found \\
\midrule
6 Choices (SCR)       & 0.493   & 0.857   & 0.998   & 0.000 \\
7 Choices (SCR)       & 0.461   & 0.814   & 0.982   & 0.000 \\
8 Choices (SCR)       & 0.427   & 0.779   & 0.916   & 0.000 \\
9 Choices (SCR)       & 0.354   & 0.625   & 0.873   & 0.000 \\
10 Choices (SCR)      & 0.329   & 0.602   & 0.848   & 0.001 \\
11 Choices (SCR)      & 0.305   & 0.571   & 0.814   & 0.001 \\
\midrule
6 Choices (PCR)       & 0.478   & 0.831   & 0.994   & 0.000 \\
7 Choices (PCR)       & 0.431   & 0.807   & 0.973   & 0.000 \\
8 Choices (PCR)       & 0.417   & 0.742   & 0.906   & 0.000 \\
9 Choices (PCR)       & 0.362   & 0.665   & 0.878   & 0.001 \\
10 Choices (PCR)      & 0.323   & 0.609   & 0.839   & 0.001 \\
11 Choices (PCR)      & 0.296   & 0.584   & 0.812   & 0.001 \\
\midrule
\bottomrule
\label{table:10choice}
\end{tabular}
\end{table}

\subsection{Examples}
To help readers better understand our tasks, we have included example inputs and outputs for each task. Please refer to Table \ref{table:7}-\ref{table:18}.

\begin{table*}[h]
\centering
\caption{SCR Example}
\begin{tabular}{|p{0.7\textwidth}|p{0.2\textwidth}|}

\toprule
Input          & Output \\
\midrule
\#\#\# Instruction:\newline
You are a legal expert who specializes in comparing user-supplied legal cases to a list of candidate legal cases, which includes titles and content. Your main function is to identify and output the title of the most similar case from the list based on the description provided. \newline
You should only output the case title and not any other information. \newline
Consider the following choices:\newline 
Choice 1: \newline
...
\newline 
Choice 2: \newline 
...
\newline 
Choice 3: \newline 
...
\newline 
Choice 4: \newline 
Case Title: BISHOP v. STEWART \newline
Date: Nov 12, 1940 \newline 
Court: District Court, E.D. Pennsylvania \newline
Judge: Hopkinson \newline
Plaintiffs: Bishop Priest, alias Lewis Johnson \newline
Defendants: Commodore Charles Stewart \newline
Case detail:
This case involved a habeas corpus petition addressed to Commodore Stewart, requiring him to produce the body of Bishop Priest, alias Lewis Johnson, who was alleged to be a minor improperly enlisted in the navy. The case focused on the legality of enlisting minors in the U.S. Navy without parental consent. The court's decision hinged on the interpretation of various acts of Congress regarding naval enlistment and the common law regarding contracts made by minors. The primary legal question was whether the enlistment of minors without parental consent was valid. Ultimately, the court found that the proof of the petitioner's minority status failed, and he was remanded. The case cited previous decisions and acts of Congress, emphasizing the importance and necessity of minors in naval service and the constitutional powers granted to Congress to raise and support a navy.
\newline 
Choice 5: \newline 
...
\newline 
Choice 6: \newline 
...
\newline 
Choice 7: \newline 
...
\newline 
Choice 8: \newline 
...
\newline 
Choice 9: \newline 
...
\newline 
Choice 10: \newline 
...           
\newline
\#\#\# Input:\newline
Case Title: UNITED STATES v. Anderson \newline
Date: Jan 1, 1869 \newline 
Court: United States Circuit Court for the District of Tennessee \newline
Judge: McNAIRY, District Judge \newline
Plaintiffs: George Bigby \newline
Defendants: Colonel Anderson \newline
Case Detail: This case involves a writ of habeas corpus directed to Colonel Anderson to bring up the body of Zebedee Bigby, who was allegedly enlisted in the army without the necessary consent from his father. The main legal questions addressed were whether the court had the right to discharge a soldier before an application to the Secretary of War proved fruitless, and whether the soldier could be discharged when the application was made not by him but by his father. The court ruled that minors enlisted without parental consent could be discharged on habeas corpus, regardless of any application to other government departments.
&\#\#\# Response: \newline
BISHOP v. STEWART. \newline   \\
\midrule
\bottomrule
\end{tabular}
\label{table:7}
\end{table*}


\begin{table*}[h]
\centering
\caption{SCR Example}
\begin{tabular}{|p{0.7\textwidth}|p{0.2\textwidth}|}

\toprule
Input          & Output \\
\midrule
\#\#\# Instruction:\newline
You are a legal expert who specializes in comparing user-supplied legal cases to a list of candidate legal cases, which includes titles and content. Your main function is to identify and output the title of the most similar case from the list based on the description provided. \newline
You should only output the case title and not any other information. \newline
Consider the following choices:\newline 
Choice 1: \newline
...
\newline 
Choice 2: \newline 
...
\newline 
Choice 3: \newline 
...
\newline 
Choice 4: \newline 
...
\newline 
Choice 5: \newline 
...
\newline 
Choice 6: \newline 
...
\newline 
Choice 7: \newline 
Case Title: Robinson v. Campbell \newline
Date: Jan 1, 1818 \newline 
Court: United States Circuit Court \newline
Judge: Mr. Justice Todd \newline
Plaintiffs: The lessor of Robinson \newline
Defendants: Campbell \newline
Case Detail: This case involves a dispute over land titles derived from grants issued by the state of Virginia, with the land falling within the boundaries of Tennessee following a boundary settlement between Virginia and Tennessee. The central legal issue was whether an equitable title (settlement rights) could be asserted as a valid claim in an action of ejectment in the circuit courts of the United States, as opposed to being a matter exclusively for equity courts. The plaintiff's title was based on a grant to John Jones dated 1787, part of which was conveyed to the plaintiff's lessor. The defendant, Campbell, claimed the land through a subsequent grant dated 1788, which was based on a settlement right previously established by a settler in 1778 and transferred to Joseph Martin. The defendant attempted to introduce evidence supporting this claim and a deed of conveyance during the pendency of the lawsuit, which could potentially impact the plaintiff's title. However, the court rejected this evidence on the basis that it was more suitable for consideration in a court of equity rather than a law court.s
\newline 
Choice 8: \newline 
...
\newline 
Choice 9: \newline 
...
\newline 
Choice 10: \newline 
...           
\newline
\#\#\# Input:\newline
Case Title: RULE S LESSEE v. PARKER \newline
Date: Feb 18, 1815 \newline 
Court: Federal Circuit Court, Nashville \newline
Plaintiffs: Rule S Lessee \newline
Defendants: Parker \newline
Case Detail: This case involves an ejectment proceeding where the title to land was contested. The land in question was originally owned by someone named Grant, who then conveyed it to the plaintiff. The property was sold for direct taxes as belonging to Grant and was purchased by Smith and Bradford. They subsequently sold it to Parker, and the collector issued a deed to Parker. A significant legal issue in this case was the adequacy of the public notices required by law before the sale. The court focused on whether all statutory requirements for the sale were met to guard against fraud and imposition, particularly the proper advertisements preceding the sale. The court found that certain procedural safeguards, such as proper advertisement and identification of Grant as the owner in the collector's book, were not sufficiently demonstrated. This was deemed fatal to the defense's case, leading to a verdict for the plaintiff.
&\#\#\# Response: \newline
Robinson v. Campbell. \newline   \\
\midrule
\bottomrule
\end{tabular}
\label{table:10}
\end{table*}

\begin{table*}[h]
\centering
\caption{SCR Example}
\begin{tabular}{|p{0.7\textwidth}|p{0.2\textwidth}|}

\toprule
Input          & Output \\
\midrule
\#\#\# Instruction:\newline
You are a legal expert who specializes in comparing user-supplied legal cases to a list of candidate legal cases, which includes titles and content. Your main function is to identify and output the title of the most similar case from the list based on the description provided. \newline
You should only output the case title and not any other information. \newline
Consider the following choices:\newline 
Choice 1: \newline
...
\newline 
Choice 2: \newline 
...
\newline 
Choice 3: \newline 
Case Title: The Philadelphia and Trenton Railroad Company vs. James Stimpson \newline 
Date: Jan 1, 1840 \newline 
Court: Circuit Court for the Eastern District of Pennsylvania \newline 
Plaintiffs: The Philadelphia and Trenton Railroad Company \newline 
Defendants: James Stimpson \newline 
Case Detail:
This case involves James Stimpson, the defendant in error, who brought an action against the Philadelphia and Trenton Railroad Company, the plaintiffs in error, for infringing his patent right. The patent, initially granted in 1831 and renewed in 1835 after the first was cancelled due to a defective specification, covered a new and useful improvement for turning short curves on railroads. At trial, significant legal discussions focused on the admissibility and validity of the renewed patent, which lacked specific recitals required by the patent act of 1832. The court ruled that the issuance of the patent under the great seal by high-ranking officials created a presumption of regularity and compliance with legal prerequisites, effectively making the patent prima facie evidence of its validity. Additionally, objections regarding prior art and the admission of evidence were addressed, with the court reinforcing strict adherence to procedural rules in patent litigation. Ultimately, the court affirmed the lower court's judgment in favor of Stimpson, emphasizing the discretion courts hold in managing procedural aspects of trials.
\newline 
Choice 4: \newline 
...
\newline 
Choice 5: \newline 
...
\newline 
Choice 6: \newline 
...
\newline 
Choice 7: \newline 
...
\newline 
Choice 8: \newline 
...
\newline 
Choice 9: \newline 
...
\newline 
Choice 10: \newline 
...           
\newline
\#\#\# Input:\newline
Case Title: Pennock v. Dialogue \newline 
Date: Oct 24, 1984 \newline 
Court: United States Supreme Court \newline 
Judge: Mr. Justice Story \newline 
Plaintiffs: Abraham L. Pennock \& James Sellers \newline 
Defendants: Adam Dialogue \newline 
Case Detail: This case involves a dispute over a patent related to an improvement in the art of making leather tubes or hose for conveying various fluids. The plaintiffs, Pennock and Sellers, who held the patent, alleged that Dialogue violated this patent. However, the original trial resulted in a verdict for Dialogue, which was upheld by the Supreme Court. The Court's decision centered on the principle that public use of an invention without the patent holder's objection signifies an abandonment of the exclusive rights later claimed through a patent. The plaintiffs' failure to secure a patent while allowing the invention to be used publicly, essentially forfeited their right to claim exclusive rights later. This case underscores the importance of timely patent protection and the implications of public use prior to patent approval.
&\#\#\# Response: \newline
The Philadelphia and Trenton Railroad Company vs. James Stimpson. \newline   \\
\midrule
\bottomrule
\end{tabular}
\label{table:10}
\end{table*}

\begin{table*}[h]
\centering
\caption{PCR Example}
\begin{tabular}{|p{0.7\textwidth}|p{0.2\textwidth}|}

\toprule
Input          & Output \\
\midrule
\#\#\# Instruction:\newline
You are a legal expert who specializes in comparing user-supplied legal cases to a list of candidate legal cases, which includes titles and content. Your main function is to identify and output the precedent case from the list based on the description provided. \newline
You should only output the reasoning process and case title. \newline
Consider the following choices:\newline 
Choice 1: \newline
...
\newline 
Choice 2: \newline 
Case Title: Clark v. Arnold \newline 
Date: Oct 1, 1803 \newline 
Court: Superior Court of North Carolina, Hillsborough \newline 
Judge: HALL, J. \newline 
Plaintiffs: Trustees of the University (Clark) \newline 
Defendants: Arnold \newline 
Case Detail:
In "Clark v. Arnold", the dispute centered on the validity of land conveyance practices and the impact of deed registration. The case involved the trustees of the University, who claimed land under confiscation acts, arguing that the land had belonged to Henry Eustace McCulloch, an absentee during the war. The key issue was whether the deed to the defendant, Arnold, which was delivered but initially taken back by the vendor (McCulloch's attorney) to secure the payment of purchase money, effectively transferred title upon its eventual registration. The court ruled that the registration of the deed related back to the time of its initial delivery, thereby confirming Arnold's title before the land could have been confiscated. This verdict upheld the principle that a registered deed, even if initially delivered conditionally, ultimately confirms the transfer of title from the delivery date, thus rejecting the plaintiffs' claims based on the confiscation acts.
\newline 
Choice 3: \newline 
...
\newline 
Choice 4: \newline 
...
\newline 
Choice 5: \newline 
...
\newline 
Choice 6: \newline 
...
\newline 
Choice 7: \newline 
...
\newline 
Choice 8: \newline 
...
\newline 
Choice 9: \newline 
...
\newline 
Choice 10: \newline 
...           
\newline
\#\#\# Input:\newline
Case Title: PATTON et al. v. REILY\newline
Date: Jan 1, 1812\newline
Court: Circuit Court, D. Tennessee\newline
Judge: M’NAIRY, District Judge\newline
Plaintiffs: PATTON et al.\newline
Defendants: REILY\newline
Case Detail:
In the case of PATTON et al. v. REILY, the court was faced with the legal intricacies of land conveyance through registration. The plaintiffs presented a deed that had not been registered at the time of the trial, arguing for its validity based on earlier proofs and registrations. The defendants challenged the legality of the deed’s evidence due to its lack of proper registration as mandated by the legislative act of 1715, which clearly required registration for the conveyance of legal estates in lands. This act substituted the common law requirement of livery of seizin with registration. The court ruled that the deed could not be admitted as evidence since it failed to meet the requisite registration requirements set by the legislation, which emphasized that a legal estate could only be passed through properly registered deeds. This decision underscored the strict adherence to legislative requirements for land conveyance, reflecting the court's stance that no legal estate passes without proper procedural compliance, as established in the precedents and statutory framework of property law.
&\#\#\# Response: \newline
Clark v. Arnold.\newline
They have precedent relation is due to their similar case details.
\newline   \\
\midrule
\bottomrule
\end{tabular}
\label{table:10}
\end{table*}

\begin{table*}[h]
\centering
\caption{PCR Example}
\begin{tabular}{|p{0.7\textwidth}|p{0.2\textwidth}|}

\toprule
Input          & Output \\
\midrule
\#\#\# Instruction:\newline
You are a legal expert who specializes in comparing user-supplied legal cases to a list of candidate legal cases, which includes titles and content. Your main function is to identify and output the precedent case from the list based on the description provided. \newline
You should only output the reasoning process and case title. \newline
Consider the following choices:\newline 
Choice 1: \newline
...
\newline 
Choice 2: \newline 
...
\newline 
Choice 3: \newline 
...
\newline 
Choice 4: \newline 
...
\newline 
Choice 5: \newline 
...
\newline 
Choice 6: \newline 
...
\newline 
Choice 7: \newline 
...
\newline 
Choice 8: \newline 
...
\newline 
Choice 9: \newline 
Case Title: Miller v. White \newline 
Date: March 29, 1934 \newline 
Court: Supreme Court of South Carolina \newline 
Judge: Justice Carter \newline 
Plaintiffs: Thomas E. Miller, Jr. \newline 
Defendants: Benjamin White \newline 
Case Detail:
This case concerns an action for specific performance of a written contract wherein the plaintiff, Thomas E. Miller, Jr., agreed to sell a lot in Charleston, SC, to the defendant, Benjamin White, for \$1,800, payable partly in cash and partly secured by a mortgage. Miller alleged that he tendered a deed to the property as per the contract, but White refused to comply with the contract terms. White's defense involved alleged defects in Miller's title, originating from prior foreclosure proceedings against other parties, which were not correctly served, potentially affecting the validity of the title Miller offered. The trial court sustained a demurrer against White's defense, leading to this appeal. The Supreme Court found that proper service in the prior foreclosure was crucial for establishing a clear title, disagreed with the trial court's dismissal of this defense, and remanded the case for further proceedings, allowing White to prove his defense related to service defects.
\newline 
Choice 10: \newline 
...           
\newline
\#\#\# Input:\newline
Case Title: Ex Parte S.C. Dept. of Social Services \newline 
Date: April 6, 1976 \newline 
Court: Supreme Court of South Carolina \newline 
Judge: Justice Littlejohn \newline 
Plaintiffs: South Carolina Department of Social Services \newline 
Defendants: Not explicitly mentioned, relates to the order concerning Terry Carter \newline 
Case Detail:
The case arose from an order by the Richland County Family Court, which directed the South Carolina Department of Social Services (D.S.S.) to fund the enrollment and maintenance of Terry Carter, a minor, at Devereaux School in Pennsylvania. Terry Carter had previously been identified as a runaway and an emotionally disturbed child requiring long-term residential treatment. The Family Court's decision aimed to provide necessary treatment for Terry at the Devereaux Schools, with expenses estimated at \$18,000 annually, which the County Department of Social Services argued was beyond their financial capacity. The South Carolina Supreme Court reversed the lower court’s order, concluding that the Family Court lacked the authority to compel the State Department of Social Services to allocate funds for Terry’s treatment outside the state, thus invalidating the directive.
&\#\#\# Response: \newline
Miller v. White\newline
They have precedent relation is because they are under the same court.
\newline   \\
\midrule
\bottomrule
\end{tabular}
\label{table:10}
\end{table*}

\begin{table*}[h]
\centering
\caption{PCR Example}
\begin{tabular}{|p{0.7\textwidth}|p{0.2\textwidth}|}

\toprule
Input          & Output \\
\midrule
\#\#\# Instruction:\newline
You are a legal expert who specializes in comparing user-supplied legal cases to a list of candidate legal cases, which includes titles and content. Your main function is to identify and output the precedent case from the list based on the description provided. \newline
You should only output the reasoning process and case title. \newline
Consider the following choices:\newline 
Choice 1: \newline
...
\newline 
Choice 2: \newline 
...
\newline 
Choice 3: \newline 
...
\newline 
Choice 4: \newline 
...
\newline 
Choice 5: \newline 
Case Title: Simms's Lessee v. Baker \newline 
Date: January 1, 1812 \newline 
Court: Circuit Court of the United States, Nashville \newline 
Judge: M'Nairy, J. \newline 
Plaintiffs: Simms's Lessee \newline 
Defendants: Baker \newline 
Case Detail:
The case involved an action of ejectment brought by Simms’s Lessee to recover possession of a tract of land granted by North Carolina. The dispute centered on the interpretation of boundary descriptions in the land grant, specifically whether the described boundaries could extend beyond explicitly stated distances to reach a natural boundary, in this case, Duck River. The plaintiff's grant started from a point where the lower line of a survey crossed Duck River, but the actual line fell short by one mile and eight poles. The court held that natural boundaries mentioned in the grant take precedence over the specified distances. Therefore, even though the physical survey fell short, the grant could extend to the river as intended, validating the plaintiff's claim up to that boundary. The decision underscored the principle that when a land grant includes a call for a natural boundary, that boundary governs, even if it contradicts specific measurements cited in the grant.
\newline 
Choice 6: \newline 
...
\newline 
Choice 7: \newline 
...
\newline 
Choice 8: \newline 
...
\newline 
Choice 9: \newline 
...
\newline 
Choice 10: \newline 
...           
\newline
\#\#\# Input:\newline
Case Title: Doss v. Tenn. Prod. Chem. Corp.\newline
Date: December 9, 1960\newline
Court: Court of Appeals of Tennessee, Middle Section\newline
Judge: Humphreys, J.\newline
Plaintiffs: Leonard Doss and Blanche Lee Doss\newline
Defendants: Tennessee Products Chemical Corporation\newline
Case Detail:
The case involved Leonard Doss and Blanche Lee Doss filing a lawsuit against the Tennessee Products Chemical Corporation for trespassing and damaging their property through tree cutting and strip mining coal. The plaintiffs claimed ownership and actual possession of the land, supported by a deed and continuous occupation over several years, sufficient to establish their claim under the relevant statutes of limitation. The defendant contested the boundaries of the property, claiming a different boundary line that would exclude their activities from the plaintiffs' land. The court found in favor of the plaintiffs, determining that the boundary line supported by longstanding recognition and actual possession was as the plaintiffs contended, not as the defendant claimed. The court held that the plaintiffs did not need to establish a valid legal title as their actual possession and the absence of valid legal title in the defendant were sufficient. The court affirmed the lower court's decision, granting an injunction against further trespass and awarding damages for the coal extracted from the property.
&\#\#\# Response: \newline
Simms's Lessee v. Baker \newline
They have precedent relation is due to their similar case details.
\newline   \\
\midrule
\bottomrule
\end{tabular}
\label{table:10}
\end{table*}


\begin{table*}[h]
\centering
\caption{LJP Example (Zero-Shot)}
\begin{tabular}{|p{0.7\textwidth}|p{0.2\textwidth}|}

\toprule
Input          & Output \\
\midrule
\#\#\# Instruction:\newline
You are a legal expert who specializes in predicting outcomes for legal cases. Utilize your internal knowledge base to predict verdict. Your main function is to anticipate the likely verdict of the legal case presented by the user.
\newline
You should only output the verdict and not any other information. \newline
Consider the following choices:\newline 
1. Defendant Wins \newline 
2. Plaintiff Wins \newline 
3. Settlement \newline 
4. Case Dismissal \newline
\newline
\#\#\# Input:\newline
Case Title: Newsom v. Newsom \newline
Date: Decided February 13, 1956 \newline
Court: Supreme Court of Mississippi \newline
Judge: Arrington, J. \newline
Plaintiffs: Mamie Newsom \newline
Defendants: Other heirs of Mrs. M.J. Newsom \newline
Case Detail: The case "Newsom v. Newsom" involved a dispute over the validity of a deed from 1922 purportedly given by Mrs. M.J. Newsom to Mamie Newsom. The primary issue was whether the deed had been validly delivered, which is critical for its effectiveness. The appellant, Mamie Newsom, argued that the deed was a valid instrument and had been properly delivered, citing a statement in a cross-bill that indicated the deed had been altered after its execution and delivery. The respondents, other heirs of Mrs. M.J. Newsom, contested the validity of the deed, arguing it had not been delivered and had been altered after execution, thereby nullifying it. They also raised issues of estoppel and laches to argue that Mamie Newsom was barred from claiming under the deed. The case included several legal debates over the interpretation of these claims and whether they could be properly introduced at different stages of the appeal process.
&\#\#\# Response: \newline
Plaintiff Wins
\newline   \\
\midrule
\bottomrule
\end{tabular}
\label{table:10}
\end{table*}


\begin{table*}[h]
\centering
\caption{LJP Example (Zero-Shot)}
\begin{tabular}{|p{0.7\textwidth}|p{0.2\textwidth}|}

\toprule
Input          & Output \\
\midrule
\#\#\# Instruction:\newline
You are a legal expert who specializes in predicting outcomes for legal cases. Utilize your internal knowledge base to predict verdict. Your main function is to anticipate the likely verdict of the legal case presented by the user.
\newline
You should only output the verdict and not any other information. \newline
Consider the following choices:\newline 
1. Defendant Wins \newline 
2. Plaintiff Wins \newline 
3. Settlement \newline 
4. Case Dismissal \newline
\newline
\#\#\# Input:\newline
Case Title: N.Y.S. Rifle \& Pistol Ass'n, Inc. v. Bruen \newline
Date: June 23, 2022 \newline
Court: Supreme Court of the United States \newline
Judge: Justice Thomas \newline
Plaintiffs: New York State Rifle \& Pistol Association, Inc., et al. \newline
Defendants: Kevin P. Bruen, in his official capacity as Superintendent of New York State Police, et al. \newline
Case Detail:
The case challenged New York's licensing regime for carrying concealed handguns, which required applicants to demonstrate a special need for self-protection distinguishable from the general community, known as the "proper cause" requirement. The petitioners argued that this standard infringed upon their Second and Fourteenth Amendment rights to bear arms, as it restricted their ability to carry handguns publicly for self-defense. This case arose in the context of New York’s long-standing regulatory approach to the public carry of handguns, starting from the early 20th century Sullivan Law to contemporary requirements that a license applicant demonstrate "good moral character" and "no good cause exists for the denial of the license." The regulation was contested on the grounds that it was not aligned with the historical tradition of firearm regulation in the United States.
&\#\#\# Response: \newline
Plaintiff Wins
\newline   \\
\midrule
\bottomrule
\end{tabular}
\label{table:10}
\end{table*}


\begin{table*}[h]
\centering
\caption{LJP Example (Zero-Shot)}
\begin{tabular}{|p{0.7\textwidth}|p{0.2\textwidth}|}

\toprule
Input          & Output \\
\midrule
\#\#\# Instruction:\newline
You are a legal expert who specializes in predicting outcomes for legal cases. Utilize your internal knowledge base to predict verdict. Your main function is to anticipate the likely verdict of the legal case presented by the user.
\newline
You should only output the verdict and not any other information. \newline
Consider the following choices:\newline 
1. Defendant Wins \newline 
2. Plaintiff Wins \newline 
3. Settlement \newline 
4. Case Dismissal \newline
\newline
\#\#\# Input:\newline
Case Title: Freeman v. South Carolina \newline
Date: October 6, 2008 \newline
Court: Supreme Court of the United States \newline
Plaintiffs: Fred Freeman \newline
Defendants: South Carolina, et al. \newline
Case Detail:
The case involves Fred Freeman, who petitioned the United States Supreme Court seeking a writ of certiorari to review a decision from the United States Court of Appeals for the Fourth Circuit. The specifics of the underlying case or the issues on appeal are not detailed in the available excerpt. Generally, a petition for a writ of certiorari is requested to appeal a lower court's decision to the Supreme Court, indicating that significant legal questions or matters of federal law were likely involved.
&\#\#\# Response: \newline
Case Dismissal
\newline   \\
\midrule
\bottomrule
\end{tabular}
\label{table:10}
\end{table*}


\begin{table*}[h]
\centering
\caption{LJP Example (Few-Shot)}
\begin{tabular}{|p{0.7\textwidth}|p{0.2\textwidth}|}

\toprule
Input          & Output \\
\midrule
\#\#\# Similar Case Example:\newline
Case Title: J. Aron \& Co. v. SemCrude, L.P.\newline
Date: June 28, 2013\newline
Court: United States Bankruptcy Court, D. Delaware\newline
Judge: Brendan Linehan Shannon\newline
Plaintiffs: J. Aron \& Company, BP Oil Supply Company, et al.\newline
Defendants: SemCrude, L.P., et al.\newline
Case Detail:
This case arose from SemCrude L.P.'s bankruptcy proceedings, where J. Aron \& Co. and other downstream purchasers filed against SemCrude and associated companies, seeking a ruling that they purchased oil and gas free of any liens despite SemCrude's financial collapse. Prior to bankruptcy, SemCrude engaged in substantial trading and midstream oil and gas services, which faltered due to massive trading losses and a subsequent liquidity crisis. The litigation addresses whether downstream purchasers, who bought oil and gas from SemCrude, did so free from claims by upstream producers who originally supplied the oil and gas. The central legal question was the applicability of liens and security interests under the U.C.C. and other state laws to the transactions made by SemCrude with the plaintiffs.\newline
Verdict: Case Dismissal\newline

\#\#\# Precedent Case Example:\newline
Case Title: Walker v. Turner\newline
Date: March 19, 1824\newline
Court: Supreme Court of the United States\newline
Judge: Justice Washington\newline
Plaintiffs: Walker\newline
Defendants: Turner\newline
Case Detail:
The case involved a land dispute where Walker, the plaintiff, sought to recover a lot in Nashville from Turner, the defendant. Walker based his claim on a deed from 1790. Turner defended his claim with a series of legal and administrative moves starting from 1804, including a sheriff's sale of the property due to a judgment for a small debt against Walker, which resulted in Turner's predecessor in title acquiring the property. This led to a series of property transfers culminating in Turner's acquisition and development of the land. Key issues revolved around the validity of the sheriff's deed and the application of Tennessee's statute of limitations regarding possession under such deeds.
\newline
Verdict: Defendant Wins\newline

\#\#\# Instruction:\newline
You are a legal expert who specializes in predicting outcomes for legal cases. Utilize your internal knowledge base to predict verdict. Your main function is to anticipate the likely verdict of the legal case presented by the user.
\newline
You should only output the verdict and not any other information. \newline
Consider the following choices:\newline 
1. Defendant Wins \newline 
2. Plaintiff Wins \newline 
3. Settlement \newline 
4. Case Dismissal \newline

\#\#\# Input:\newline
Case Title: MOORE v. BROWN ET AL\newline
Date: January 1, 1850\newline
Court: U.S. Supreme Court\newline
Plaintiffs: Joshua J. Moore\newline
Defendants: James Brown, Alfred Brown, Harmon Hogan, and Joseph Froward\newline
Case Detail:
The case centered around a deed issued by the Illinois Auditor of Public Accounts, purportedly under authority to sell land for unpaid taxes as per an 1823 act. The deed was challenged on the basis that it violated statutory requirements, notably because the sale occurred earlier than permitted by law. The plaintiffs argued that the deed, showing a sale date that did not comply with the mandatory notice period, was void and could not confer title to the defendants. This raised questions about the application of the Illinois statute of limitations, specifically whether a deed void on its face due to procedural defects could support a defense of adverse possession under a color of title.
&\#\#\# Response: \newline
Defendant Wins
\newline   \\
\midrule
\bottomrule
\end{tabular}
\label{table:10}
\end{table*}


\begin{table*}[h]
\centering
\caption{LJP Example (Few-Shot)}
\begin{tabular}{|p{0.7\textwidth}|p{0.2\textwidth}|}

\toprule
Input          & Output \\
\midrule
\#\#\# Similar Case Example:\newline
Case Title: M'Donald's Heirs v. Smalley\newline
Date: January 1, 1832\newline
Court: Supreme Court of the United States\newline
Judge: Chief Justice Marshall\newline
Plaintiffs: M'Donald's Heirs\newline
Defendants: Smalley\newline
Case Detail:
The case involved a dispute over land ownership in Ohio, where M'Donald's Heirs sought to secure land that was held by Smalley under a senior patent. The plaintiffs claimed the land based on a prior entry made in the name of David Anderson, who was deceased at the time of the entry. This prior entry was crucial as it formed the foundation of the plaintiffs' claim. The case centered on whether an entry made in the name of a deceased person could be valid, a point that had previously been addressed in another case, Galt et al. v. Galloway.
\newline
Verdict: Case Dismissal\newline

\#\#\# Precedent Case Example:\newline
Case Title: De La Vergne Refrigerating Machine Co. v. Featherstone\newline
Date: Decided January 9, 1893\newline
Court: United States Supreme Court\newline
Judge: Chief Justice Fuller\newline
Plaintiffs: De La Vergne Refrigerating Machine Co.\newline
Defendants: Featherstone et al.\newline
Case Detail:
The case centered around the validity of a patent issued after the death of the inventor, James Boyle. The patent was issued to Boyle, "his heirs or assigns," which raised questions about its validity since Boyle had died before the patent was granted. This led to a discussion on whether the patent could be validly issued to his legal representatives or heirs under existing patent laws. The case delved into whether the administrative process followed by Boyle's legal representatives, including filing amendments and maintaining the application posthumously, adhered to patent laws and whether such actions could legitimately sustain the patent's validity.
\newline
Verdict: Plaintiff Wins\newline

\#\#\# Instruction:\newline
You are a legal expert who specializes in predicting outcomes for legal cases. Utilize your internal knowledge base to predict verdict. Your main function is to anticipate the likely verdict of the legal case presented by the user.
\newline
You should only output the verdict and not any other information. \newline
Consider the following choices:\newline 
1. Defendant Wins \newline 
2. Plaintiff Wins \newline 
3. Settlement \newline 
4. Case Dismissal \newline

\#\#\# Input:\newline
Case Title: Atlas T. W. Mufflers v. McCallum\newline
Date: January 23, 1929\newline
Court: Supreme Court of Texas\newline
Judge: Justice Speer\newline
Plaintiffs: Atlas Trailers Water Mufflers, Inc.\newline
Defendants: Mrs. McCallum, Secretary of State\newline
Case Detail:
This case involved Atlas Trailers Water Mufflers, Inc., seeking a writ of mandamus to compel the Secretary of State, Mrs. McCallum, to file the company's amended charter, which included patents valued at \$50,000 as capital stock. The Secretary of State had denied the filing based on a long-standing departmental policy that did not recognize patents as tangible property suitable for capitalizing a corporation. The company argued that patents, being property capable of assignment and possessing an ascertainable value, should be recognized as valid contributions towards the capital stock under the relevant Texas constitutional and statutory provisions.
&\#\#\# Response: \newline
Plaintiff Wins
\newline   \\
\midrule
\bottomrule
\end{tabular}
\label{table:10}
\end{table*}


\begin{table*}[h]
\centering
\caption{LJP Example (Few-Shot)}
\begin{tabular}{|p{0.7\textwidth}|p{0.2\textwidth}|}

\toprule
Input          & Output \\
\midrule
\#\#\# Similar Case Example:\newline
Case Title: Stewart v. Griffith\newline
Date: April 25, 1910\newline
Court: Supreme Court of the United States\newline
Judge: Justice Holmes\newline
Plaintiffs: The executor of one Ball (Stewart)\newline
Defendants: Griffith\newline
Case Detail:
This case centers on a dispute over a contract for the sale of real estate, where the executor of a deceased's estate seeks specific performance of a contract made by the appellant to purchase land. The contract had provisions that were to result in forfeiture and make the contract null and void if certain conditions were not met. The key issue was whether these conditions allowed the appellant to withdraw from the contract or obligated him to complete the purchase as per the initial agreement. The executor argued that despite the death of the property owner just before the finalization of the sale, the contractual obligations still stood, entitling the estate to enforce the contract. The complexities of the case involve interpretations of Maryland real estate law, the powers of an executor under a will, and the legal implications of contract terms that stipulate conditions for forfeiture and nullification.
\newline
Verdict: Defendant Wins\newline
\#\#\# Precedent Case Example:\newline
Case Title: Willis v. First Real Estate Investment Co.\newline
Date: January 24, 1934\newline
Court: Circuit Court of Appeals, Fifth Circuit\newline
Judge: Hutcheson, Circuit Judge\newline
Plaintiffs: Henry B. Willis\newline
Defendants: First Real Estate Investment Company and others\newline
Case Detail:
This case involves a property dispute where Henry B. Willis challenged the validity of land titles held by the First Real Estate Investment Company and others, based on historical claims. The conflict arises from a Mexican title originating in 1927 and a Texas title from 1861. Willis's claim is grounded in the assertion that changes in the river's course—specifically the avulsive changes referenced in boundary treaties—affected the jurisdiction over the land, which was located along the Texas bank of the Rio Grande. The case examines intricate historical and legal arguments surrounding land ownership, jurisdictional changes due to natural river movements, and the implications of international treaties between the U.S. and Mexico.
\newline
Verdict: Defendant Wins\newline
\#\#\# Instruction:\newline
You are a legal expert who specializes in predicting outcomes for legal cases. Utilize your internal knowledge base to predict verdict. Your main function is to anticipate the likely verdict of the legal case presented by the user.
\newline
You should only output the verdict and not any other information. \newline
Consider the following choices:\newline 
1. Defendant Wins \newline 
2. Plaintiff Wins \newline 
3. Settlement \newline 
4. Case Dismissal \newline

\#\#\# Input:\newline
Case Title: San Lorenzo T. I. Co. v. City Mortgage Co.\newline
Date: June 30, 1934\newline
Court: Supreme Court of Texas\newline
Judge: Justice Pierson\newline
Plaintiffs: San Lorenzo Title and Improvement Company\newline
Defendants: City Mortgage Company\newline
Case Detail:
The case involves a trespass to try title suit regarding land along the Rio Grande, designated as a "banco" under treaties between the USA and Mexico. The San Lorenzo Title and Improvement Company claimed title to the land, arguing it derived from Mexican governmental and court actions before the International Boundary Commission declared the land a banco in 1930 and stated it belonged to the USA. The core of the dispute rested on the effect of the 1905 treaty which aimed to resolve the issues of bancos along the Rio Grande by stipulating those on the north bank would pass to the USA. The company contended that prior Mexican claims to the land should be recognized despite the treaty's provisions.
&\#\#\# Response: \newline
Defendant Wins
\newline   \\
\midrule
\bottomrule
\end{tabular}
\label{table:18}
\end{table*}

\end{document}